\documentclass[10pt,twocolumn,letterpaper]{article}

\usepackage[pagenumbers]{cvpr} % To force page numbers, e.g. for an arXiv version

\usepackage{graphicx}
\usepackage{amsmath}
\usepackage{amssymb}
\usepackage{booktabs}

\usepackage[pagebackref,breaklinks,colorlinks]{hyperref}

\usepackage[capitalize]{cleveref}
\crefname{section}{Sec.}{Secs.}
\Crefname{section}{Section}{Sections}
\Crefname{table}{Table}{Tables}
\crefname{table}{Tab.}{Tabs.}

\def\cvprPaperID{2169} % *** Enter the CVPR Paper ID here
\def\confName{CVPR}
\def\confYear{2023}

\usepackage{pifont}
\usepackage[ruled,vlined]{algorithm2e}
\usepackage[dvipsnames]{xcolor}
\usepackage{array}
\usepackage{booktabs}
\usepackage{subcaption}
\usepackage{enumitem}

\usepackage{soul}
\usepackage{comment}

\def\B#1{\textbf{#1}}
\def\BR#1{[#1]}
\def\BB#1{[\B{#1}]}
\def\IT#1{\textit{#1}}
\newcommand{\cmark}{\ding{51}}%

\def \su#1{\small{$^#1$}\large}

\begin{document}

%%%%%%%%% TITLE
\title{Weakly-Supervised Text-driven Contrastive Learning for Facial Behavior Understanding}

% \author{Xiang Zhang, Taoyue Wang, Xiaotian Li, Lijun Yin\\
% \and
% Huiyuan Yang\\

\author{\parbox{16cm}{\centering
    {\large Xiang Zhang\su{1}\hspace{5mm} Taoyue Wang\su{1}\hspace{5mm} Xiaotian Li\su{1}\hspace{5mm} Huiyuan Yang\su{2}\hspace{5mm} Lijun Yin\su{1}}\\
    \vspace{2mm}
    {\large \hspace{6mm}\su{1}State University of New York at Binghamton} \hspace{8mm} 
    {\large \su{2}Rice University} \\
    {\tt\small \{zxiang4, twang61, xli210, lyin\}@binghamton.edu} \hspace{5mm} 
    {\tt\small hy48@rice.edu}
    % \thanks{This work was not supported by any organization}% <-this % stops a space
}
}

\maketitle

%%%%%%%%% ABSTRACT
\begin{abstract}
Contrastive learning has shown promising potential for learning robust representations by utilizing unlabeled data. 
However, constructing effective positive-negative pairs for contrastive learning on facial behavior datasets remains challenging. 
This is because such pairs inevitably encode the subject-ID information, and the randomly constructed pairs may push similar facial images away due to the limited number of subjects in facial behavior datasets.
To address this issue, we propose to utilize activity descriptions, coarse-grained information provided in some datasets, which can provide high-level semantic information about the image sequences but is often neglected in previous studies.
% By leveraging such activity information, effective positive-negative pairs is created for contrastive learning.
More specifically, we introduce a two-stage \B{C}ontrastive \B{L}earning with Text-\B{E}mbeded framework for \B{F}acial behavior understanding (\B{CLEF}).
The first stage is a weakly-supervised contrastive learning method that learns representations from positive-negative pairs constructed using coarse-grained activity information.
The second stage aims to train the recognition of facial expressions or facial action units by maximizing the similarity between the image and the corresponding text label names.
The proposed \B{CLEF} achieves state-of-the-art performance on three in-the-lab datasets for AU recognition and three in-the-wild datasets for facial expression recognition.
\end{abstract}

%%%%%%%%% BODY TEXT

\begin{figure}[th]
  \centering
  \begin{subfigure}[t]{0.99\linewidth}
    \includegraphics[width=\textwidth]{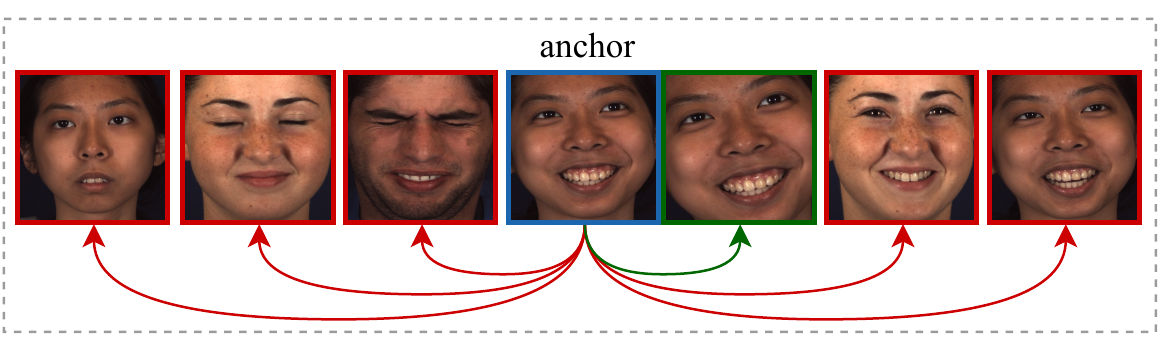}
    \caption{Self-supervised contrastive learning pairs}
  \end{subfigure}
  \hfill
  \begin{subfigure}[t]{0.99\linewidth}
    \includegraphics[width=\textwidth]{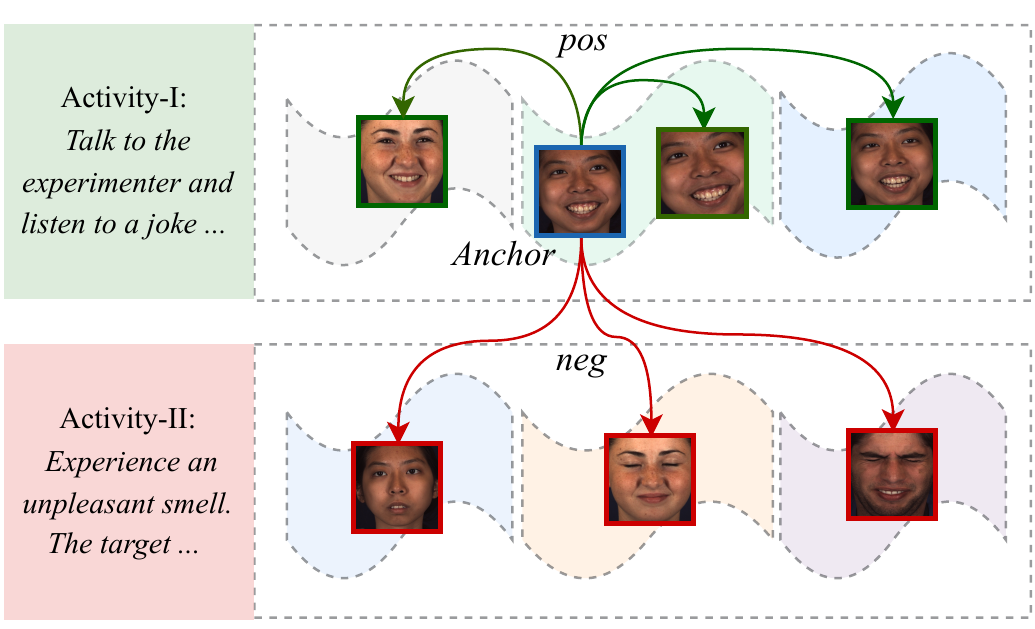}
    \caption{Activity-based weakly-supervised contrastive learning pairs}
    \label{fig:figure_pre2}
  \end{subfigure}
  \caption{(a) shows the self-supervised contrastive learning paring, where green represents positive pairs and red represents negative pairs. 
  In a batch, the only positive samples for an anchor are its augmentations, while all others are negative. Even if the last image is similar (same person and same expression) to the anchor, it will be pushed away from the anchor as a negative sample.
  (b) is the illustration of the proposed weakly-supervised contrastive learning method: 
    samples from the same activity in a batch are selected as positive and the remaining are negative.
    The textual activity descriptions are used as coarse-grained information to guide contrastive learning, for example, ``talk to the experimenter and listen to a joke ..."
    }
  \label{fig:figure_pre}
\end{figure}

% \begin{figure}[t]
%   \centering
%     \includegraphics[width=0.48\textwidth]{figures/figure_pre_sub1}
%     \caption{Illustration of the proposed weakly-supervised Contrastive Learning Method. 
%     Anchor (blue) and its variations from the same activity are selected as positive samples (green) while images from the other activities are selected as negative samples (red). 
%     Activity descriptions are also split into positive-negative sets by the same rule.
%     We leverage the activity classes as coarse-grained labels for weakly-supervised representation learning.}
%     \label{fig:figure_pre}
% \end{figure}

%------------------------------- Introduction ------------------------------------------
\section{Introduction}
\label{sec:intro}
Facial expression is one of the most natural signals to analyze human emotion and behavior.
Ekman~\cite{ekman1971} has indicated that facial expressions of emotion are universal across human cultures and categorized them, apart from neutral expression, into six categories: anger, disgust, fear, happiness, sadness, and surprise. 
Then contempt was added as another basic emotion, according to the work~\cite{matsumoto1992}.
Furthermore, facial expressions are coded by specific facial muscle movements, called Action Units (AUs) in Facial Action Coding System (FACS)~\cite{ekman1997}.
Automatic Facial Expression Recognition (FER) and Action Unit recognition (AUR) have been core problems in facial analysis, attracting significant interest in the computer vision community.

Recently, many deep learning-based approaches~\cite{liu2015spontaneous, yang2018, ruan2020deep, dmue2021, sev,faut,ksrl} have been proposed and achieved state-of-the-art performance in FER and AUR.
A variety of methods~\cite{yang2018, ran2020, eac, zhang2018identity} aimed to disentangle the expression or AU features from various disturbing factors, such as identity, ethnic background, pose, etc.
Along with the development of Self-Supervised Learning (SSL), unlabeled data is utilized for learning good representations to improve recognition performance.
Chang et al.~\cite{ksrl} proposed a rule that divides the face into eight regions, which are then fed in a contrastive learning component.
Shu et al.~\cite{shu2022revisiting} explored three core strategies in self-supervised contrastive learning to enforce expression-specific representations and minimize interference from other facial attributes.
FaRL~\cite{farl} proposed a vision-language pre-training model with a large number of facial image-text pairs to learn facial representation.
To build an appropriate self-supervised learning task, fine-grained auxiliary information, such as landmarks and image captions, is typically required, which in turn requires more data processing.

On the other hand, several works have investigated the different relations between AU pairs and their applications.
SRERL~\cite{srerl} was developed to learn the appearance representation of the semantic relationships between AUs by a graph convolutional network.
Yang et al.~\cite{sev} proposed a cross-modal attention module to enhance the image representations by including AU semantic descriptions.
However, due to the low consistency between the data structure of image and text, attention-based integration may not fully exploit the potential of textual data.
Some works also modeled the AUs' relationships with the expressions to improve the FER performance.
Cui et al.~\cite{cui2020knowledge} employed a Bayesian Network(BN) to capture the generic knowledge on relationships among AUs and expression.
In our work, we are interested in learning the direct relationships between expressions and between AUs in a simpler way.
Moreover, previous studies on relationship learning have rarely explored the representation of ground truth labels, instead focusing on fitting the model with numerical labels, thus sparking our interest in investigating label representation.

In order to overcome the above limitations, it is necessary to investigate the following two issues:   
\noindent\textit{i) whether there is any coarse-grained information, which can be \B{easily obtained and simple to use} without compromising the performance}; 
\noindent\textit{ii) whether there is any approach to \B{enrich the relationship information } of the label representation}.

% 
% Hence, we propose a method based on contrastive learning, which learns the similarity of the same class across different modalities.

To address the above two issues, we propose a text-driven contrastive learning method, called CLEF, to utilize both the coarse-grained information and text-embedded labels.
The proposed method comprises two stages, both using a unified vision-text architecture known as CLIP~\cite{clip}.
In pre-training, for each anchor in a batch, we consider positive samples from the same activity and negative samples from different activities.
The activity descriptions are used as coarse-grained labels to guide the weakly-supervised contrastive learning model that aims to minimize the intra-activity differences in representations.
Table~\ref{tab:activity} shows some samples of activity descriptions of BP4D~\cite{bp4d}.
Figure~\ref{fig:figure_pre2} shows how we leverage the activity descriptions to create positive-negative pairs.
Each activity contains multiple expressions, but our pairing construction can increase the possibility of grouping images with the same expression into positive ones.
The distance between images belonging to different activities increases, even if the images have the same identities, which encourages the encoder to focus on the activity features rather than the identity features.
% Note that the person in the left image of Activity-II in Figure~\ref{fig:figure_pre2} is the same person as in the anchor, but it is pushed away from the anchor, which reduces the impact of the identity features.
Meanwhile, the activity text description does not contain any identity information, allowing the text encoder to avoid encoding identity features.
Cross-modal contrastive learning is therefore designed to push image features close to such textual features. 
% The core idea of this selection of sample pairs is to mitigate the effects of personal attributes such as identity, gender, age, and ethnic background.
Performing on these pairs can enhance the learning of better representations, which in turn improves the performance of FER or AUR in downstream tasks.

In fine-tuning, we apply vision-text contrastive learning directly to classification tasks. 
Supervised contrastive learning adapts the image representation to be close to its corresponding label name feature, while self-supervised contrastive loss encourages the feature of label names and descriptions to be similar, enriching the semantic information of the label representation.
Therefore, we believe such label representation is more powerful than the numerical label.
The recognition prediction is based on finding the most similar label names of the testing image, following the method used in CLIP~\cite{clip}.
The main contributions of this paper are summarized in three aspects:

% %%%%%%%%%%%%%%%%%%%%%%%   Activity Table ####################################

\begin{table}[t]
\caption{Activity description samples in BP4D. See more descriptions in the Supplementary Material.}
\vspace*{-1em}
\label{tab:activity}
\begin{center} %
\renewcommand{\arraystretch}{1.0}
\begin{tabular}{|l|m{7.0cm}|}
\hline
        & Activity Description   \\
\hline
A1       & Talk to the experimenter and listen to a joke (Interview). The target emotion is happiness or amusement \\ 
\hline
A2       & Watch and listen to a recorded documentary and discuss their reactions. The target emotion is sadness  \\
\hline
A3       & Experience sudden, unexpected burst of sound. The target emotion is surprise or startle \\
\hline
A4       & Play a game in which they improvise a silly song. The target emotion is embarrassment \\
\hline
% A5       & Anticipate and experience physical threat. The target emotion is fear or nervous \\
% \hline
% A6       & Submerge their hand in ice water for as long as possible. The target emotion is physical pain \\
% \hline
% A7       & Experience harsh insults from the experimenter. The target emotion is anger or upset \\
% \hline
% A8       & Experience an unpleasant smell. The target emotion is disgust \\
% \hline
\end{tabular}
% \vspace*{-1em}
\end{center}
\vspace{-1.5em}
\end{table}

\begin{enumerate}[topsep=-1pt,itemsep=0.2ex,partopsep=1ex,parsep=1ex]
\item We proposed a weakly-supervised contrastive learning method that effectively leverages coarse-grained activity information. 
It not only requires less data processing but also learns better representations.

\item We explore the use of text-driven contrastive learning on FER and AUR tasks, where the performance is improved by incorporating textual information.

\item Extensive experiments have been conducted on 3 in-the-lab datasets and 3 in-the-wild datasets. 
The proposed method achieves state-of-the-art performance in all 6 datasets, demonstrating the effectiveness of the proposed method.
 
\end{enumerate}

% \begin{figure}[t]
%   \centering
%     \includegraphics[width=0.48\textwidth]{figures/figure_fine.pdf}
%     \caption{Supervised Contrastive Learning on downstream facial expression recognition task.}
%     \label{fig:short-a}
% \end{figure}

%------------------------------- Related Works ------------------------------------------
\section{Related works}

\subsection{Facial Expression Recognition}
In order to improve the performance of facial expression recognition, various deep neural networks are designed with different insights on FER to obtain powerful representations.
Researchers have conducted a series of studies~\cite{yang2018, liu2019hard, ruan2020deep, fdrl2021} aiming to decompose different attributes from facial behavioral representations and learn robust expression-related features.
Another line of methods has aimed to enhance intra-class compactness and reduce inter-class compactness in feature extractions~\cite{li2017reliable, cai2018island}. 
Additionally, several works explore the attention mechanism on FER to obtain the discriminative features~\cite{ran2020,li2020attention,li2021your}.
Furthermore, multi-task learning has been employed in various approaches, including involving facial landmark learning~\cite{devries2014multi}, AU recognition ~\cite{kollias2019face,cui2020knowledge}, and others.
Recently, due to the successful recognition performance on laboratory databases, more researchers attempt to perform FER model on in-the-wild databases, which typically contain significant label noises.
As a result, addressing such noisy label issues has become a popular topic in the recent research community, as evidenced by several works~\cite{scn2020, dmue2021, rul2021, eac2022}.

\subsection{Facial Action Uniti Recognition}

In recent years, deep learning has been applied to facial action unit recognition, leading to significant improvements in performance.
Some works have focused on learning better facial features by emphasizing important local regions, also known as regions of interest (ROI)~\cite{drml, eac, jaa}.
Considering the interdependency between different AUs, several works have applied graph neural networks (GCN) to model these relations~\cite{song2021uncertain, song2021hybrid, srerl, anfl}.
Recent works involved multiple techniques to improve the recognition accuracy, including transformer methods~\cite{faut}, self-supervised methods~\cite{ksrl}, and semi-supervised methods~\cite{piap}.
Focus on the input data, some recent work~\cite{amf,mft,li2023disagreement} utilized multi-modal learning methods with other modalities, such as depth images, and thermal images.

SEV-Net~\cite{sev} is the first work that exploited semantic text-embedding of AU description on AUR, where the AU relationships are learned by these descriptions.
The cross-modal attention mechanism between semantic embeddings and image features is used to enhance the discriminative features.
\B{Instead of} cross-modal attention, we employ text-driven contrastive learning to enhance image-text features, which then improves performance on both FER and AUR.

\subsection{Contrastive Learning}

Recently, We have witnessed the potential of contrastive learning in representation learning.
The principle of contrast learning is to make positive sample pairs consistent and negative sample pairs exclusive.
It has been widely applied to unsupervised learning works~\cite{chen2020simple,he2020momentum,chen2021exploring} with outstanding success in representation learning.
SupCon~\cite{khosla2020supervised} extends contrastive learning to a fully supervised setting, named Supervised contrastive learning.
In this work, data belonging to the same class are selected as positive samples, and data from different classes as negative.

\B{Text-driven Recognition.} 
Text-driven recognition has become an active area in both Natural Language Processing (NLP) and Computer Vision (CV).
In this area, common tasks include visual question answering~\cite{antol2015vqa}, image captioning~\cite{vinyals2015show}, and image-text retrieval~\cite{chen2020uniter}.
Pioneering work CLIP~\cite{clip} not only demonstrates that image-text contrastive learning achieves promising performance for visual representation learning but also brings textual supervision into the classic recognition tasks in CV.
Researchers have extended this vision-language model to other areas, such as object detection~\cite{gu2021open}, image segmentation~\cite{li2021language}, and video action recognition~\cite{wang2021actionclip}.
Recent FaRL~\cite{farl} explores this vision-language model on facial representation learning by pre-training on a variety of facial image-text pairs. 
However, only the image encoder was evaluated on several downstream tasks.
\B{In contrast}, CLEF utilizes the text encoder in downstream facial behavior analysis tasks, resulting in better performance than using only the image encoder.

%------------------------------- Methodology ------------------------------------------

\begin{figure}[!h]
\centering
\includegraphics[width=0.47\textwidth]{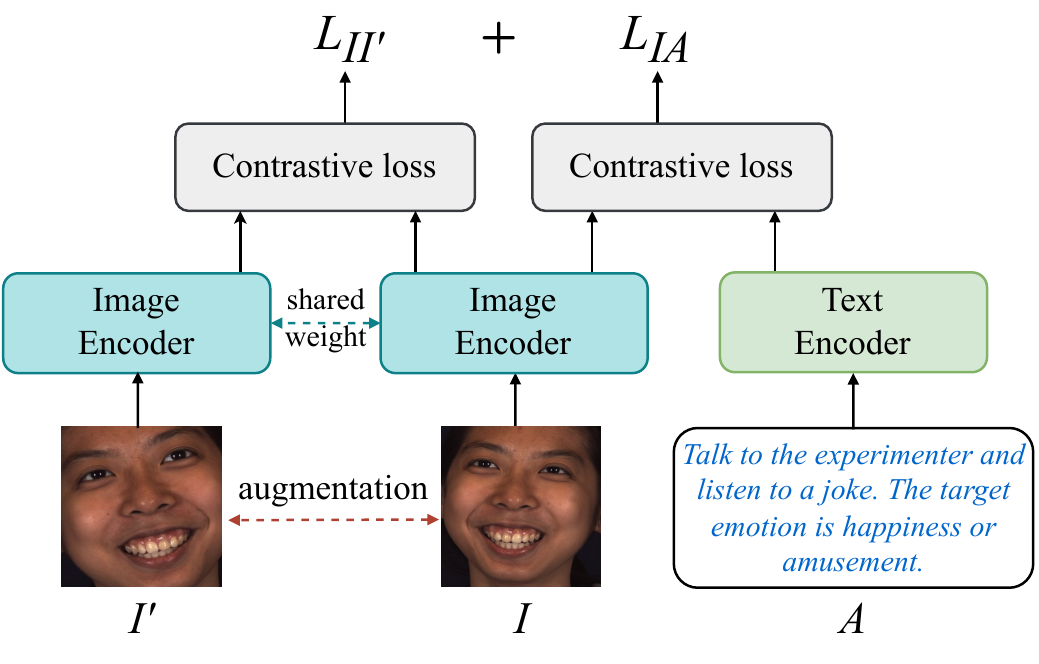}
\vspace{-.5em}
\caption{An overview of the architecture of the proposed CLEF in pre-training.
$\mathcal{L}_{II'}$ and $\mathcal{L}_{IA}$ indicate supervised contrastive loss between images and between images and activity descriptions, respectively.
}
\vspace{-0.9em}
\label{fig:arch_pre}
\end{figure}

\begin{figure*}
\centering
\includegraphics[width=0.98\textwidth]{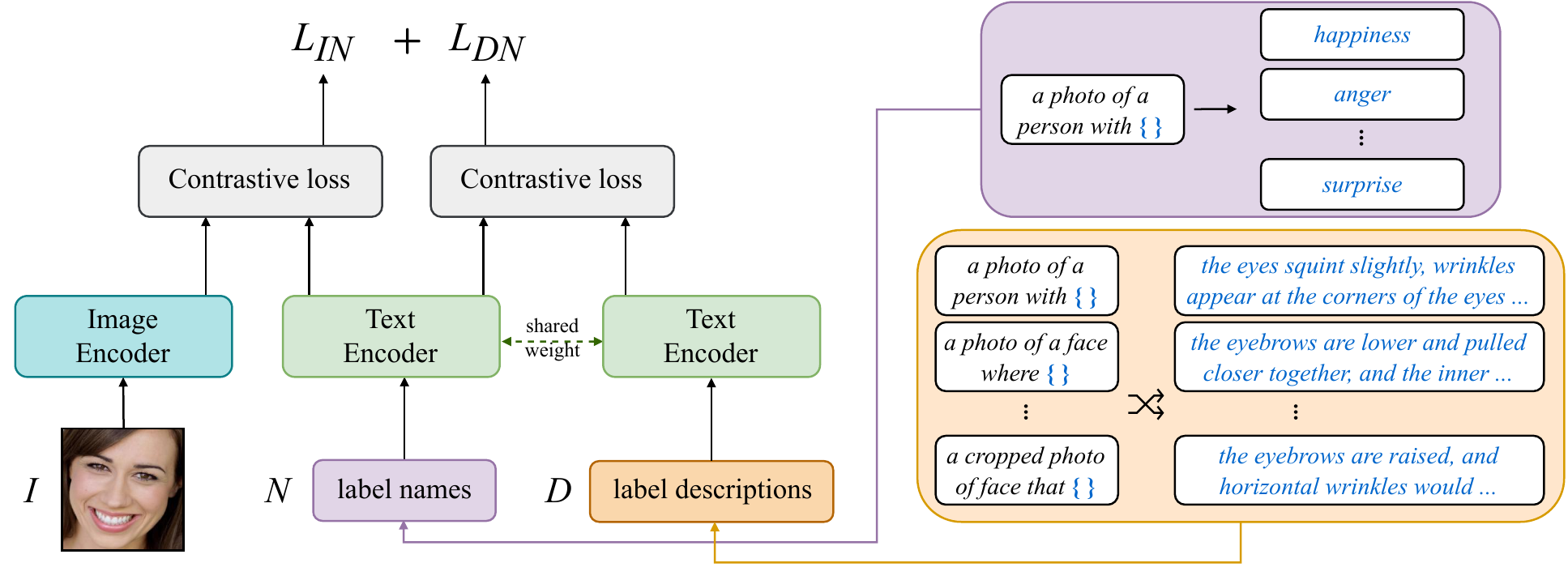}
\vspace{-.5em}
\caption{An overview of the architecture of the proposed CLEF in the downstream FER task. 
It is based on the CLIP model which consists of an image encoder and a text encoder.
$\mathcal{L}_{IN}$ indicates supervised contrastive loss between images and label names.
The self-supervised contrastive loss $\mathcal{L}_{DN}$ between label descriptions and label names is jointly adapted.}
\label{fig:arch_fine}
\end{figure*}

\section{Methodology}

Our proposed framework consists of two stages, and each is built with an image-text encoder, the same as CLIP~\cite{clip}.
Figure~\ref{fig:arch_pre} and Figure~\ref{fig:arch_fine} show the overview of architectures in pre-training and fine-tuning respectively. 

\subsection{Pre-training}
In pre-training, we aim to learn robust deep representation and alleviate the influence of identity variation. 
Therefore, we designed a contrastive learning task that pulls together features from the same activity and pushes them away from features of other activities.
Sets of images, activities, and activity labels are defined by $I$, $A$, and $Y^A$.
Given $n$ samples in a mini-batch, we generate two augmentations, $X^I$, and $\tilde{X}^I$.
The extracted feature representations by image encoder $g(\cdot)$ and text encoder $h(\cdot)$ are $z^I=g(x^I)$, $\tilde{z}^I=g(\tilde{x}^I)$ , and $z^A=h(x^A)$ where $x^I \in X^I$, $\tilde{x}^I \in \tilde{X}^I$, and $x^A \in A$.
Meanwhile, the labels are duplicated to $2n$ as $\tilde{Y}^A$.
Inspired by the work~\cite{khosla2020supervised}, we consider images and their corresponding textual activity descriptions under the same activity as positive samples.
We propose the cross-modal supervised contrastive loss and leverage the coarse-grained activity label to guide contrastive learning in pre-training.

\B{Cross-modal Supervised Contrastive Loss}: 
The contrastive loss, in the scenario of $z$, $y$ pairs, at temperature $\epsilon$ is defined as:

\begin{align}
  \mathcal{L}_{\alpha\beta}^\text{sup} = -\sum_{i=1}^{n}\frac{1}{2 N_i} \sum\limits_{j \in J} \log \frac{\exp(\mathbf{z}^\alpha_i \cdot \mathbf{z}^{\alpha+\beta}_j / \epsilon)}{\sum\limits_{k \in K} \exp({\mathbf{z}^\alpha_i \cdot \mathbf{z}^{\alpha+\beta}_k / \epsilon})}
\end{align}
where the symbol $(\cdot)$ denotes the inner (dot) product, $J \equiv 2N(y^A_i), j \neq i$; 
$K \equiv \{i\}^{2n}_{i=1}, k \neq i$, and $\alpha$, $\beta$ are from the extracted multi-modal feature sets.
$N_i= \{j \in \{i\}^{n}_{i=1}: \tilde{y}^A_j = \tilde{y}^A_i \}$ contains a set of indices of positive samples with label $y^A_i$.

Given the $z^{I\tilde{I}} \in \{Z^I, \tilde{Z}^I\}$, $z^{IA} \in \{Z^I, Z^A\}$, the final loss in pre-training is:
\begin{align}
    \mathcal{L}_{pre} = \mathcal{L}_{I\tilde{I}}^\text{sup} + \mathcal{L}_{IA}^\text{sup}
\end{align}

$\mathcal{L}_{I\tilde{I}}^\text{sup}$ encourages similar representations for images from the same activity, while $\mathcal{L}_{IA}^\text{sup}$ encourages similar representations between images and their corresponding texts.
Given that similar facial behaviors are more likely to appear within the same activity, the encoder tends to focus on capturing facial behavior features while avoiding personal attributes features, such as identity, gender, and ethnicity.
% Thus, our weakly-supervised method can learn similar representations for data in the same activity and vice versa.

\subsection{Fine-tuning}

Unlike the previous work~\cite{farl} which only utilized the image encoder in downstream facial analysis tasks, our approach is in the scenarios of both image and text, as we believe that text contains useful information for facial behavior analysis.
% In this stage, we only apply one image augmentation for classification.
Given a set of images, label names, and label descriptions. i.e., $I$, $N$, $D$.
Similar to the pre-training, the extracted features representations are $\{z^I, z^N, z^D\}$ by image encoder $g(\cdot)$ and text encoder $h(\cdot)$.
Our loss functions include self-supervised contrastive loss and supervised contrastive loss for name-description and image-label pairs respectively.
The self-supervised contrastive loss, in the scenario of name-description pairs, is given as
\begin{align}
    \mathcal{L}_{DN} = - \frac{1}{C} \sum_{i=1}^{C} \log \frac{\exp(\mathbf{z}^D_i \cdot \mathbf{z}^N_j / \tau)}{\sum_{j=1}^{C} \exp(\mathbf{z}^D_i \cdot \mathbf{z}^N_j / \tau)} 
    \label{eq:loss_dl}
\end{align}
where $C$ is the class number, e.g., 12 AUs, and 8 expressions. $\tau$ is a learnable parameter of the temperature to scale the logits.

The supervised-contrastive learning for image-text pairs is jointly trained. We design different supervised contrastive losses based on cross-entropy and binary cross-entropy loss, as FER is a multi-class classification problem and AUR is a multi-label problem.
In the FER task, the loss is defined as,
\begin{align}
    \mathcal{L}_{IN}^{fe} = - \frac{1}{B} \sum_{i=1}^{B} \sum_{c=1}^{C} w_{c}\log \frac{\exp(\mathbf{z}^I_i \cdot \mathbf{z}^N_c / \tau) y_i^c}{\sum_{j=1}^{C} \exp(\mathbf{z}^I_i \cdot \mathbf{z}^N_j / \tau)} 
    \label{eq:loss_il_fe}
\end{align}
where $B$ is the batch size, $w$ is the weight, $y$ is the target of ground truth.

The loss in AUR is formulated as,
\begin{align}
    \mathcal{L}_{IN}^{au} = - \frac{1}{B}\sum_{i=1}^{B} \sum_{c=1}^{C}(w_c y_i^c log(\sigma(\mathbf{z}^I_i \cdot \mathbf{z}^N_c / \tau)) + \nonumber \\
    (1-y_i^c) log(1-\sigma(\mathbf{z}^I_i \cdot \mathbf{z}^N_c / \tau)))
    \label{eq:loss_il_au}
\end{align}

where $\sigma$ is the activation function $sigmoid(\cdot)$.

Consequently, the total loss function in fine-tuning is defined as:
\begin{align}
    \mathcal{L}_{fine} = (\lambda \mathcal{L}^{\alpha}_{IN} + \mathcal{L}_{DN}) / 2.0
    \label{eq:loss_fine}
\end{align}

where the $\alpha\in\{ fe, au\}$, $\lambda$ is a hyperparameter.

The loss function $\mathcal{L}_{IN}$ forces the image features to be close to the target textual features of label names.
The self-supervised contrastive loss $\mathcal{L}_{DN}$ leverages the inter-class difference of semantic information to enhance the feature extracted from the text encoder.
By jointly training both self-supervised and supervised contrastive components, our method learns not only the inter-class relations but also features in shared latent space across modalities, where the textual feature is unbiased to every subject identity.
The algorithms' pseudocodes in PyTorch-style are shown in the Supplementary Material.

\subsection{Text Prompting}
Following CLIP~\cite{clip}, we also use prompt templates to augment the original label in our method. 
% We only use one prompt template ``a photo of a person with \{label name\}" for label names, and randomly select one template for label description and activity description in a mini-batch.
% The semantic description is written according to the interpretation of previous psychological works~\cite{ekman1971,matsumoto1992,ekman1997}.
% Examples of label name and label descriptions are shown in Figure~\ref{fig:arch_fine}.
% The detail of prompting is in the Supplementary Material.
We only use one prompt template ``a photo of a person with \{label name\}." for label names, e.g., ``a photo of a person with \IT{happiness}.", "a photo of a person with \IT{inner brow raiser}.".
For label descriptions, we prepare multiple prompt templates on them, e.g., ``a photo shows a person that \{label description\}.", ``a cropped photo of face that \{label description\}.''. 
Considering the limited number of label descriptions in databases, instead of ensembling all prompt templates by their mean textual representation, we randomly select one prompt template in training. 
Similarly, activity descriptions are also randomly applied with prompt templates, e.g., ``a photo of an activity that \{activity description\}.'', ``a photo of a person from an activity that \{activity description\}.'' 
The detail of prompting is in the Supplementary Material.

%------------------------------- Experiments ------------------------------------------

\section{Experiments}

The proposed CLEF is compared with the state-of-the-art methods on six popular databases for FER and AUR tasks.
Furthermore, we conduct ablation studies to verify the component-wise contribution of our method.

\subsection{Databases}

\subsubsection{AU Databases}

\hspace{4mm}\B{BP4D}~\cite{bp4d} contains 41 subjects captured in laboratory environments.
There are 8 activities designed to elicit different spontaneous emotions, resulting in $41 \times 8$ video clips.
Expert coders select the most expressive 20 seconds of each video clip for AU coding, producing 140,000 labeled frames.
Following the work~\cite{eac}, we split all labeled frames into subject-exclusive 3-fold with 12 AUs for both two stages.

\B{BP4D+}~\cite{bp4d+}
consists of 140 subjects with a total of 1.5 M frames in the same laboratory environments.
For each subject, 20 seconds from 4 activities are annotated, resulting in 192,000 labeled frames.
First, the 140 subjects are split into four-fold, following the same setting in ~\cite{mft}. 
In pre-training, we equally sample 480,000 frames from all 1.5 M frames by 10 activity categories.
In fine-tuning, 12 AUs, the same as in BP4D, are selected for AU recognition.

\B{DISFA}~\cite{disfa}
contains videos from the left view and right view of 27 subjects. 
In the same manner as ~\cite{sev}, we choose 8 of 12 AUs with AU intensities higher or equal to 2 as positive samples.
The model trained on BP4D is then fine-tuned to the DISFA dataset, which is following the setting in ~\cite{eac, srerl}.
F1-score is reported based on subject-exclusive 3-fold cross-validation.

\subsubsection{FE Databases}

\hspace{4mm}\B{AffectNet}~\cite{affectnet} is currently the largest FER dataset, including 440,000 images with manual annotation of 8 basic expressions. 
AffectNet-7 refers to a manually annotated set without contempt class, resulting in 283,901 and 3,500 images for training and testing respectively.
AffectNet-8 includes all expression images with 287,568 training samples and 4,000 testing samples.

\B{RAF-DB}~\cite{raf_db} is labeled by 15,000 facial images with 7 expressions, i.e., neutral, happiness, surprise, sadness, anger, disgust, and fear. 
Following the previous work setting~\cite{dmue2021}, we choose 12,271 images for training and the remaining 3,068 for testing. 

\B{FERPlus}~\cite{fer_plus} is an extended version of FER2013~\cite{fer2013}, where 8 emotions (with contempt) are annotated. It contains 28,709 training images, 3,589 validation images, and the remaining 3,589 testing images. 
For a fair comparison, we report the accuracy on the test set with the same setting from~\cite{ran2020}.

         %%%%%% Table 1 Single Modality on BP4D %%%%%%

\begin{table*}[hbt]
\caption{F1 scores in terms of 12 AUs on BP4D. Bold numbers indicate the best performance; bracketed numbers indicate the second best.}
\label{tab:result_bp4d}
\centering %
\renewcommand{\arraystretch}{1.}
\begin{tabular}{l|*{12}{>{\centering\arraybackslash}p{20pt}}|c}
\toprule
Methods             & AU1       & AU2       & AU4       & AU6       & AU7       & AU10      & AU12      & AU14      & AU15      & AU17      & AU23      & AU24      & Avg\\
\toprule
EAC \cite{eac}      & 39.0      & 35.2      & 48.6      & 76.1      & 72.9      & 81.9      & 86.2      & 58.8      & 37.5      & 59.1      & 35.9      & 35.8      & 55.9\\
DSIN\cite{dsin}     & 51.7      & 40.4      & 56.0      & 76.1      & 73.5      & 79.9      & 85.4      & 62.7      & 37.3      & 62.9      & 38.8      & 41.6      & 58.9\\
JAA-Net\cite{jaa}   & 47.2      & 44.0      & 54.9      & 77.5      & 74.6      & 84.0      & 86.9      & 61.9      & 43.6      & 60.3      & 42.7      & 41.9      & 60.0\\
HMP-PS\cite{hmpps}  & 53.1      & 46.1      & 56.0      & 76.5      & 76.9      & 82.1      & 86.4      & 64.8      & 51.5      & 63.0      & 49.9      & 54.5      & 63.4\\
SEV-Net\cite{sev}   & \B{58.2}  & \B{50.4}  & 58.3      & \B{81.9}  & 73.9      & \B{87.8}  & 87.5      & 61.6      & 52.6      & 62.2      & 44.6      & 47.6      & 63.9\\
FAUT\cite{faut}     & 51.7      & 49.3      & \BR{61.0} & 77.8      & 79.5      & 82.9      & 86.3      & \BR{67.6} & 51.9      & 63.0      & 43.7      & \BR{56.3} & 64.2\\
PIAP\cite{piap}     & 55.0      & \BR{50.3} & 51.2      & \BR{80.0} & 79.7      & 84.7      & \B{90.1}  & 65.6      & 51.4      & \BR{63.8} & \BR{50.5} & 50.9      & 64.4\\
KSRL\cite{ksrl}     & 53.3      & 47.4      & 56.2      & 79.4      & \B{80.7}  & 85.1      & 89.0      & 67.4      & \B{55.9}  & 61.9      & 48.5      & 49.0      & 64.5\\
ANFL~\cite{anfl}    & 52.7      & 44.3      & 60.9      & 79.9      & \BR{80.1} & \BR{85.3} & \BR{89.2} & \B{69.4}  & \BR{55.4} & \B{64.4}  & 49.8      & 55.1      & \BR{65.5}\\
\midrule
\B{CLEF}            & \BR{55.8} & 46.8      & \B{63.3}  & 79.5      & 77.6      & 83.6      & 87.8      & 67.3      & 55.2      & 63.5      & \B{53.0}  & \B{57.8}  & \B{65.9}\\
\bottomrule
\end{tabular}

\end{table*}

                %%%%%% Table 2 Single Modality on DISFA %%%%%%

\begin{table*}[ht]
\caption{F1 scores in terms of 8 AUs on DISFA. Bold numbers indicate the best performance; bracketed number indicate the second best.}
\label{tab:result_disfa}
\centering %
\renewcommand{\arraystretch}{1.}
\begin{tabular}{l|*{8}{>{\centering\arraybackslash}p{1.3cm}}|c}
\toprule
Methods             & AU1       & AU2       & AU4       & AU6       & AU9       & AU12      & AU25      & AU26      & Avg\\
\toprule
% DRML\cite{drml}     & 17.3      & 17.7      & 37.4      & 29.0      & 10.7      & 37.7      & 38.5      & 20.1      & 26.7\\
EAC \cite{eac}      & 41.5      & 26.4      & 66.4      & 50.7      & \B{80.5} & \B{89.3} & 88.9      & 15.6      & 48.5\\
DSIN \cite{dsin}    & 42.4      & 39.0      & 68.4      & 28.6      & 46.8      & 70.8      & 90.4      & 42.2      & 53.6\\
JAA-Net\cite{jaa}   & 43.7      & 46.2      & 56.0      & 41.4      & 44.7      & 69.6      & 88.3      & 58.4      & 56.0\\
HMP-PS\cite{hmpps}  & 38.0      & 45.9      & 65.2      & 50.9      & 50.8      & 76.0      & 93.3      & \B{67.6} & 61.0\\
SEV-Net\cite{sev}   & 55.3      & 53.1      & 61.5      & [53.6]      & 38.2      & 71.6      & \B{95.7} & 41.5      & 58.8\\
FAUT\cite{faut}     & 46.1      & 48.6      & \B{72.8}  & \B{56.7} & 50.0      & 72.1      & 90.8      & 55.4      & 61.5\\
PIAP\cite{piap}     & 50.2      & 51.8      & [71.9]      & 50.6      & 54.5      & \BR{79.7}  & \BR{94.1}  & 57.2      & 63.8\\
KSRL\cite{ksrl}     & \BR{60.4}  & \BR{59.2}  & 67.5      & 52.7      & 51.5      & 76.1      & 91.3      & \BR{57.7}  & \BR{64.5} \\
ANFL~\cite{anfl}    & 54.6      & 47.1      & \BB{72.9} & \B{54.0}  & \B{55.7}  & 76.7      & 91.1      & 53.0      & 63.1\\
\midrule
% \B{CLEF-zero}\\
% \B{CLEF-few}\\
\B{CLEF}            & \B{64.3} & \B{61.8}	& 68.4	    & 49.0	    & [55.2]	    & 72.9	    & 89.9	    & 57.0	    & \B{64.8}\\
\bottomrule
\end{tabular}

\end{table*}

\subsection{Implementation Details}

\B{Model Architecture.}
The proposed model consists of a text encoder $h(\cdot)$ of transformer~\cite{transformer} model, and an image encoder $g(\cdot)$ of ViT~\cite{vit} model to learn textual features and visual features respectively. 
Specifically, the image encoder is ViT-B/16 with 12-layer and 768-width, resulting in 87M parameters with the input of $3 \times 224 \times 224$. 
The input image is first split into 14 × 14 patches, and then 14 × 14 patch embeddings are obtained by linear projection. 
A learnable cls token is inserted at the beginning of these embeddings, and then we can get 197 embeddings by adding position embeddings. 
The text encoder is a 12-layer, 512-width, and 8-head Transformer with 63M parameters. 
The length of the input text token is 77, and truncation or padding is performed if the input length does not match.
We project features from both the image cls token and the text eos token to 512 widths as the output logits.
Finally, we calculate the contrastive losses by the normalized output logits.

\B{Pre-training setup.}
BP4D and BP4D+ contain the activity descriptions for our weakly-supervised contrastive learning in the first stage.
Model parameters are loaded from FaRL~\cite{farl} during this stage.
mage augmentation techniques such as random cropping, horizontal flipping, and random rotation are used.
We set the batch size by 64 and choose Adamw~\cite{adamw_2019} optimizer with 0.01 weight decay.
The model has been trained 5 epochs with 1 epoch warmup, followed by cosine decay~\cite{sgdr_2016} with a minimal learning rate of 1.e-6.
The fixed temperature $\epsilon$ is set at 0.25.

\B{Downstream tasks setup.}
In downstream fine-tuning, lr of $2\times10^{-4}$ is set in BP4D, AffectNet, RAF-DB and $10^{-4}$ in DISFA and FER+.
The model is trained with 64 batch-size and an Adamw optimizer.
The evaluation metric for AUR is the averaged F1-score over all AUs, and for FER it is accuracy. 
Hyperparameter $\lambda$ is set to 2 and its investigation is in the Supplementary Material.
Other implementation details can also be found in the Supplementary Material.

%%%%%% Single Modality on BP4D+ %%%%%%
\begin{table*}[t!]
\caption{F1 scores in terms of 12 AUs on BP4D+. Bold numbers indicate the best performance; bracketed numbers indicate the second best.}
\label{tab:result_bp4d+}
\centering %
\renewcommand{\arraystretch}{1.0}
\begin{tabular}{l|*{12}{>{\centering\arraybackslash}p{19pt}}|c}
\toprule
Methods             & AU1       & AU2       & AU4       & AU6       & AU7       & AU10      & AU12      & AU14      & AU15      & AU17      & AU23      & AU24      & Avg\\
\toprule
ViT~\cite{vit}      & 45.6      & 38.2      & 35.5      & 85.9      & 88.3      & 90.3      & [89.0]    & 81.9      & 45.8      & 48.8      & 57.2      & 34.6      & 61.6\\
CLIP~\cite{clip}    & \B{49.4}  & [39.7]    & [38.9]    & 85.7      & 87.6      & [90.6]    & [89.0]    & 80.6      & 44.9      & 50.3      & 56.1      & 32.8      & 62.1\\
EAC \cite{eac}      & 43.7      & 39.0      & 14.0      & 85.6      & 87.2      & 90.5      & 88.7      & \B{88.4}  & 45.7      & 49.0      & [57.3]    & \B{43.6}  & 61.1\\
JAA \cite{jaa}      & 46.0	    & \B{41.3}  & 36.0      & 86.5	    & [88.5]	& 90.5	    & 89.6      & 81.1	    & 43.4	    & 51.0	    & 56.0	    & 32.6	    & 61.9\\
SEV-Net \cite{sev}  & 47.9      & 40.8      & 31.2      & \B{86.9}  & 87.5      & 89.7      & 88.9      & [82.6]    & 39.9      & \B{55.6}  & \B{59.4}  & 27.1      & 61.5 \\
MFT \cite{mft}      & [48.4]    & 37.1	    & 34.4	    & 85.6	    & \B{88.6}  & \B{90.7}  & 88.8      & 81.0	    & \B{47.6}  & [51.5]	& 55.6	    & 36.9	    & [62.2]\\
\midrule
\B{CLEF}            & 47.5	    & 39.6	    & \B{40.2}	& [86.5]	& 87.3	    & 90.5	    & \B{89.9}	& 81.6	    & [47.0]	& 46.6	    & 54.3	    & [41.5]	& \B{63.1}\\
\bottomrule
\end{tabular}
\end{table*}

\subsection{Comparison with the State of the Art}

\subsubsection{Facial Action Unit recognition}

We compare our method with several state-of-the-art works, namely EAC~\cite{eac}, DSIN~\cite{dsin}, JAA-Net~\cite{jaa}, HMP-PS~\cite{hmpps}, SEV-Net~\cite{sev}, FAUT~\cite{faut}, PIAP~\cite{piap}, KSRL~\cite{ksrl} and ANFL~\cite{anfl} on BP4D and DISFA datasets. 
Table~\ref{tab:result_bp4d} shows the comparison result on BP4D in terms of the F1-score of 12 AUs.
Overall, CLEF achieves outstanding performance on the widely used database and outperforms the state-of-the-art methods in 3 AUs, namely AU4, AU23, and AU24. 
In addition, the quantitative results on the DISFA database are reported in Table~\ref{tab:result_disfa}, where CLEF achieves the best performance on average F1-score in terms of 8 AUs.

Table~\ref{tab:result_bp4d+} shows the comparison results of our proposed method CLEF with ViT~\cite{vit}, CLIP~\cite{clip}, EAC~\cite{eac}, JAA~\cite{jaa}, SEV-Net~\cite{sev}, and MFT~\cite{mft} on the BP4D+ database.
ViT and CLIP are used as the baseline methods, while the results of EAC and JAA are reported in the work of MFT.
Our method performs better than the state-of-the-art methods in terms of 12 AUs, with an overall improvement of 1.4\%.

\subsubsection{Facial Expression Recognition}
To demonstrate the generalization ability of CLEF, we also conduct experiments on the facial expression recognition task.
The performance of CLEF is evaluated on the facial expression recognition task, and the results are shown in Table~\ref{tab:result_fer} on three commonly used in-the-wild FER databases.
The state-of-the-art works are including RAN~\cite{ran2020}, SCN~\cite{scn2020}, RUL~\cite{rul2021}, DMUE~\cite{dmue2021}, VTFF~\cite{VTFF2021} and the most recent EAC~\cite{eac2022}.
The model is fine-tuned from the pre-trained CLEF on BP4D+.
Our method achieves the best performance than other state-of-the-art methods on AffectNet-7, RAF-DB, and FER+, while slightly lower than DMUE under AffectNet-8.

%%%%%%%%%%%%%%%%%%%%%%   FER Result Table ####################################

\begin{table}[th]
\caption{Facial expression recognition accuracies on 3 FER databases. AN-7: AffectNet-7, AN-8: AffectNet-8.
Bold numbers indicate the best performance; racketed numbers indicate the second best.}

\label{tab:result_fer}
\centering %
\renewcommand{\arraystretch}{1.0}
\begin{tabular}{l|*{4}{c}}
\toprule
Methods                 & AN-7       & AN-8       & RAF-DB       & FER+  \\
\toprule
RAN \cite{ran2020}      & 59.50      & -          & 86.90        & 88.55    \\
SCN \cite{scn2020}      & 63.40	     & 60.23      & 87.03        & 88.01	\\
RUL \cite{rul2021}      & 61.43      & -          & 88.98        & 88.75    \\
DMUE \cite{dmue2021}    & -          & \B{62.84}  & 88.76        & 88.64    \\
VTFF \cite{VTFF2021}    & 64.80      & 61.85      & 88.14        & 88.81    \\
EAC \cite{eac2022}      & [65.32]    & -          & [89.99]      & [89.64] \\
\midrule
\B{CLEF}                & \B{65.66}	 & [62.77]	  & \B{90.09}	 & \B{89.74} \\
\bottomrule
\end{tabular}
\vspace*{-1em}
\end{table}

\subsection{Zero-shot Evaluation}

We evaluate our model using zero-shot settings, where training a model with Neutral, Happiness, and Fear on AffectNet and test it by Sadness, Surprise, Disgust, and Anger on RAF-DB and FER+.
See the results in the left part of Table~\ref{tab:result_zeroshot}.
Additionally, we also evaluated FER on all expressions using a BP4D+ AUR model, shown in the right section of Table~\ref{tab:result_zeroshot}.
Label descriptions are used to infer the model.
Since the model is unaware of the unseen label names, label descriptions are used in inference.
Zero-shot is challenging, but CLEF outperforms the baseline FaRL obviously.

%%%%%%%%%%%%%%%%%%%%%%% Zero-shot Recognition ####################################

\begin{table}[h!]
\caption{Zero-shot results on RAF-DB and FER+}
\vspace{-10pt}
\label{tab:result_zeroshot}
\begin{center} %
\renewcommand{\arraystretch}{1.0}
\begin{tabular}{l|cc|cc}    
\hline
Methods     & RAF-DB      & FER+        & RAF-DB    & FER+     \\ % 'AU9' 'AU25', 'AU26'
\hline
FaRL        & 16.21       & 25.73       & 13.10     & 21.20  \\ 
\hline
CLEF        & \B{29.14} & \B{34.40}          & \B{29.47}      & \B{24.90} \\ 
\hline
\end{tabular}
\end{center}
\end{table}

\subsection{Ablation Study}
% \vspace{-5pt}

%%%%%%%%%%%%%%%%%%%%%%% Ablation Study Table ####################################

\begin{table}[ht]
\caption{Evaluation of key components on BP4d and RAF-DB. Results indicate F1-score on BP4D, while accuracy on RAF-DB. PA: pre-trained with activity texts. PI: pre-trained with image. I: image encoder. N: label names. D: label descriptions.}
\vspace*{-.5em}
\label{tab:result_ablation}
\centering %
\renewcommand{\arraystretch}{1.0}
\begin{tabular}{l*{5}{c}|*{2}{c}}
\toprule
Methods     & PA    & PI     & I      & N      & D      & BP4D      & RAF-DB \\
\toprule
CLIP        &  	     &  	     & \cmark &        &        & 63.4      & 87.88\\
CLIP        &  	     &  	     & \cmark & \cmark &        & 64.0      & 88.72\\
CLIP        &  	     &  	     & \cmark & \cmark & \cmark & 64.4      & 89.70\\
\midrule
FaRL        &  	     &  	     & \cmark &        &        & 63.7      & 88.31\\
FaRL        &  	     &  	     & \cmark & \cmark &        & 64.1      & 88.69\\
FaRL        &  	     &  	     & \cmark & \cmark & \cmark & 64.6      & 88.78\\
\midrule
CLEF        &        & \cmark   & \cmark &        &        & 65.0	    & 89.67 \\
CLEF        & \cmark & \cmark   & \cmark &        &        & 64.2	    & 89.34 \\
CLEF        &        & \cmark   & \cmark & \cmark &        & 64.7       & 88.57 \\
CLEF        & \cmark & \cmark   & \cmark & \cmark &        & 64.9       & 89.57 \\
CLEF        &        & \cmark   & \cmark & \cmark & \cmark & 64.8       & 89.44\\
CLEF        & \cmark & \cmark   & \cmark &        & \cmark & 65.7       & 89.73\\
CLEF        & \cmark & \cmark   & \cmark & \cmark & \cmark & \B{65.9}   & \B{90.09}\\

\bottomrule
\end{tabular}
\vspace*{-1em}
\end{table}

% \B{Evaluation of different $\lambda$ and text mask ratio.}
% In this section, we evaluate the performance on BP4D by setting different hyperparameters $\lambda$ in Equation~\ref{eq:loss_fine}.
% Moreover, different mask ratios of mask tokens in the label description input is also investigated in Figure~\ref{fig:lambada_mask_ratio}.

% Moreover, to clarify the effectiveness of self-supervised contrastive learning between DN, we also compare our method with another baseline method, which is also involving D.
% We design the baseline method by loss function $\mathcal{L}_{IN} + \mathcal{L}_{ID}$, where the model is trained on the similarity between I and N, and also between I and D.
% We test this loss function on BP4D, where the baseline function achieves at in the average F1-score in terms of 12 AUs.

% \subsection{Zero-shot classification evaluation}

To evaluate the effectiveness of each component in CLEF, we conducted ablation studies on both AUR and FER tasks.
We assessed the contributions of each important component in our method, i.e., pre-trained stage with images (PI), pre-trained stage with activity texts (PA), image encoder (I), label names (N), and label description (D). 
It is worth noting that the text encoder is trainable only when N or D is available.
Otherwise, the image feature is followed by a linear projection as the output for supervised learning. 
N and D are also two modalities that contribute to the contrastive losses in Equations~\ref{eq:loss_dl},~\ref{eq:loss_il_fe},~\ref{eq:loss_il_au}.
Table~\ref{tab:result_ablation} shows the performance of various combinations of the components.
The original CLIP and FaRL are used as the baseline methods for comparison.
The result shows our model effectively learns features in the pre-training stage and leads to an improvement in recognition performance.
Specifically, using the image encoder alone in pre-training (PI) results in some improvement (65.0 on BP4D), adding the textual activity (PA) and text encoder (ND) further improves the performance (65.9 on BP4D). 
Additionally, regardless of pre-training, a model with the text encoder using N and D achieves better performance than a single image encoder.

Contrastive learning between names and descriptions not only enhances the text feature from names but also expands the distinction among different descriptions.
If names such as `Disgust', and `Fear' are isolated points in a high dimensional space, descriptions such as `...eyebrows are pulled down...'  and `...eyebrows are pulled up...' are more likely to be surfaces interacted at specific points.
Hence, when utilizing contrastive learning, the distance between the corresponding name-description becomes closer, while the distance between inter-descriptions is also further.
The best performance is achieved by using both names and descriptions, which demonstrated that there's an optimal balance between `distinction' and `similarity'.

\B{Weight-shared Text Encoder} 
We share the weight of the text-encoder to extract the features of label names and label descriptions respectively in fine-tuning.
We assume label names and label descriptions are projected in the same features space, where the distance depends on words combinations;
Otherwise, the contrastive learning of relationships is limited by cross-spaces.
Meanwhile, feeding the names and descriptions into different text encoders could reduce the input diversities, which can lead to performance degradation. 

\begin{figure}[th]
  \centering
  \begin{subfigure}{0.49\linewidth}
    \includegraphics[width=\textwidth]{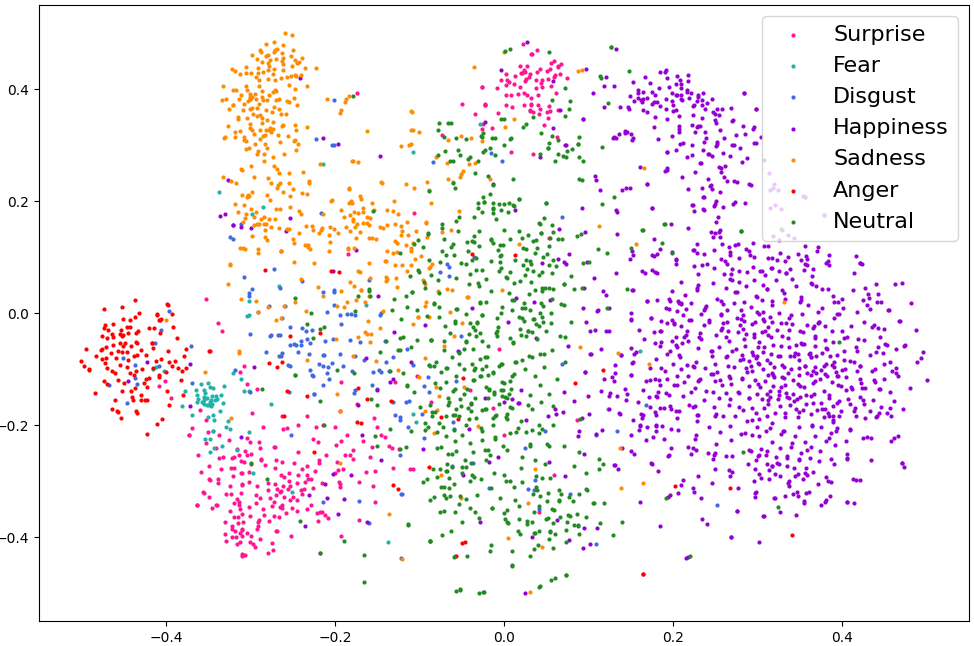}
    \caption{Baseline}
  \end{subfigure}
  \hfill
  \begin{subfigure}{0.49\linewidth}
    \includegraphics[width=\textwidth]{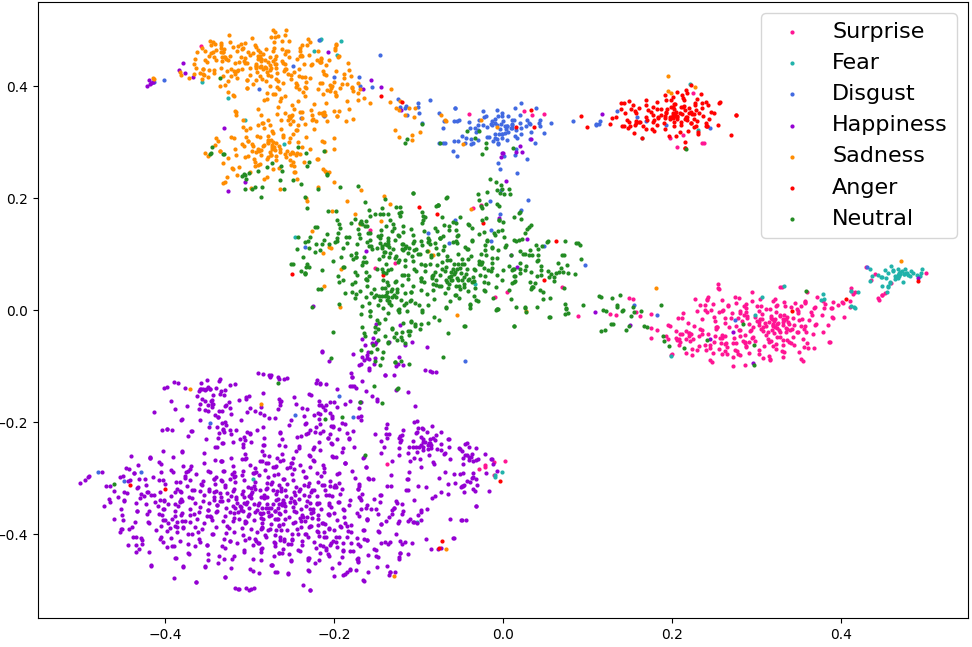}
    \caption{CLEF}
  \end{subfigure}
  \caption{t-SNE visualization of the expression features on RAF-DB. }
  \label{fig:tsne}
\end{figure}

\begin{figure}[th]
  \centering
  \begin{subfigure}[t]{0.49\linewidth}
    \includegraphics[width=\textwidth]{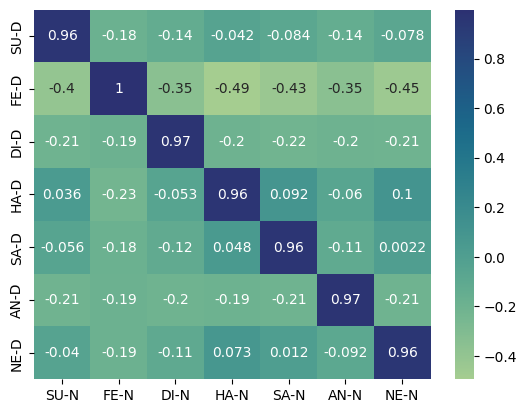}
    \caption{Cosine similarity matrix between features of label descriptions and label names trained on RAF-DB}
  \end{subfigure}
  \hfill
  \begin{subfigure}[t]{0.495\linewidth}
    \includegraphics[width=\textwidth]{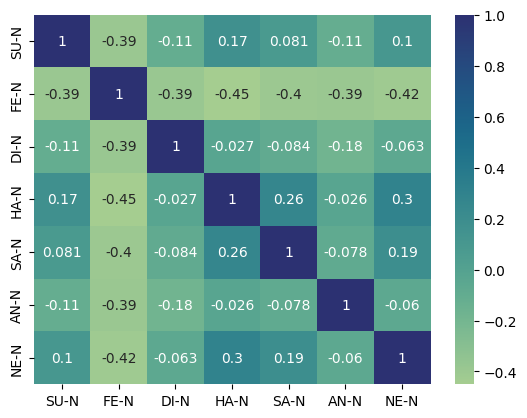}
    \caption{Pearson correlation coefficient matrix of label name features trained on RAF-DB}
  \end{subfigure}
  \caption{Visualization of similarities on RAF-DB. SU: Surprise, FE: Fear, DI: Disgust, HA: Happiness, SA: Sadness, AN: Anger, NE: Neutral. D: description, N: name}
  \label{fig:matrix}
\end{figure}

Such an assumption means our model not only reduces the model size but also achieves better performance than the application of two separated text encoders.
Hence, we continue to conduct experiments based on two individual text encoders on BP4D, achieving the average F1-score of 64.5, which is worse than using weight-shared text encoders. 

\subsection{Visualization}

Figure~\ref{fig:tsne} shows t-SNE~\cite{tsne} visualization of visual expression features extracted by the baseline method (FaRL) and the proposed CLEF on RAF-DB, respectively. 
The expression features extracted by the baseline method are not easily distinguishable from different facial expressions, while the proposed CLEF effectively enhances the separability of different classes.
In particular, CLEF makes the differences among neutral, disgust, and sadness more pronounced compared to the baseline.
We visualized the similarity matrix and correlation coefficient matrix of the text features on RAF-DB, which is shown in Figure~\ref{fig:matrix}.

\begin{figure*}[t]
  \centering
  \begin{subfigure}{0.55\linewidth}
    \includegraphics[width=\textwidth]{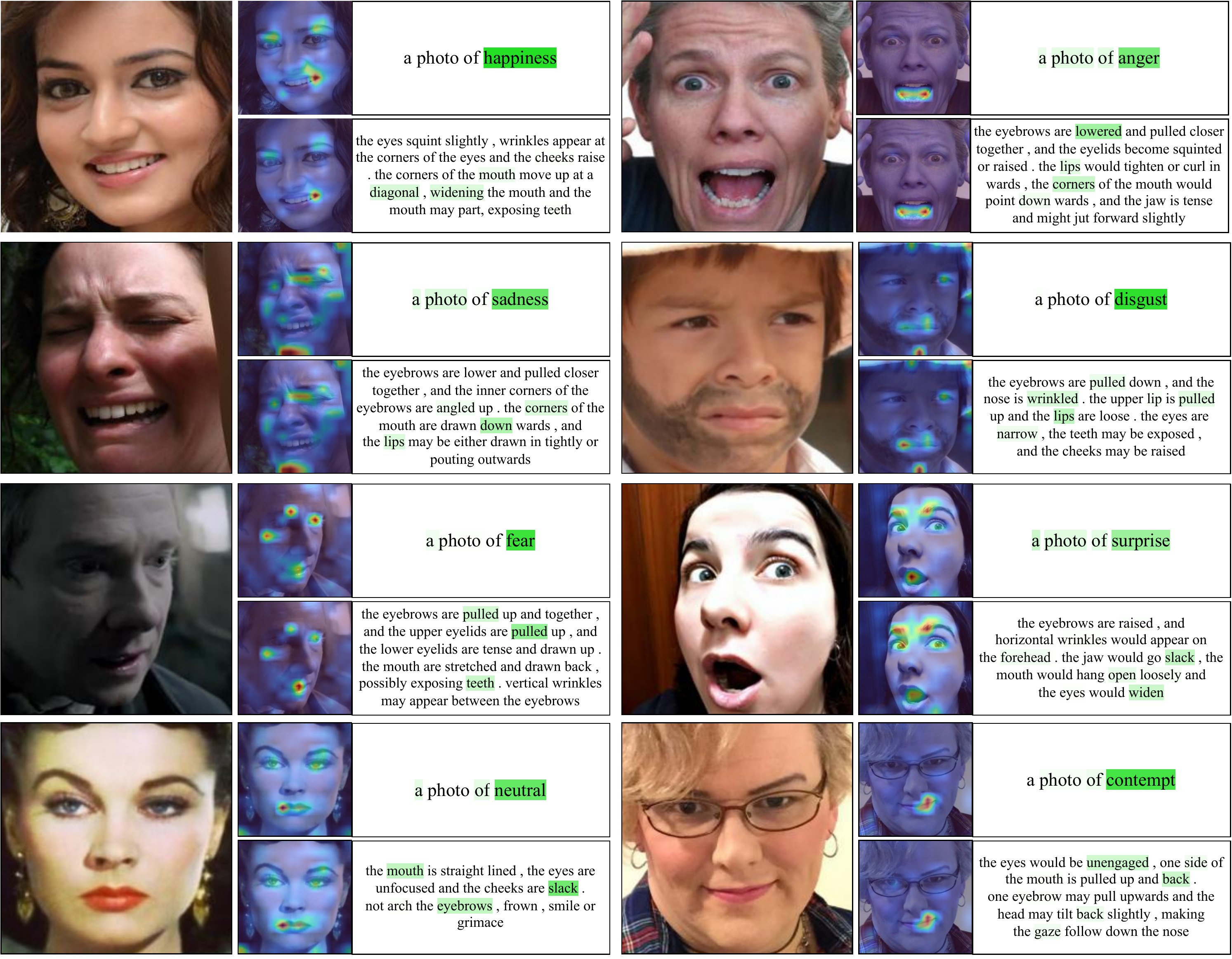}
    \caption{Heatmap samples of 8 expressions on AffectNet}
  \label{fig:heatmap_fer}
  \end{subfigure}
  \hfill
  \begin{subfigure}{0.438\linewidth}
    \includegraphics[width=\textwidth]{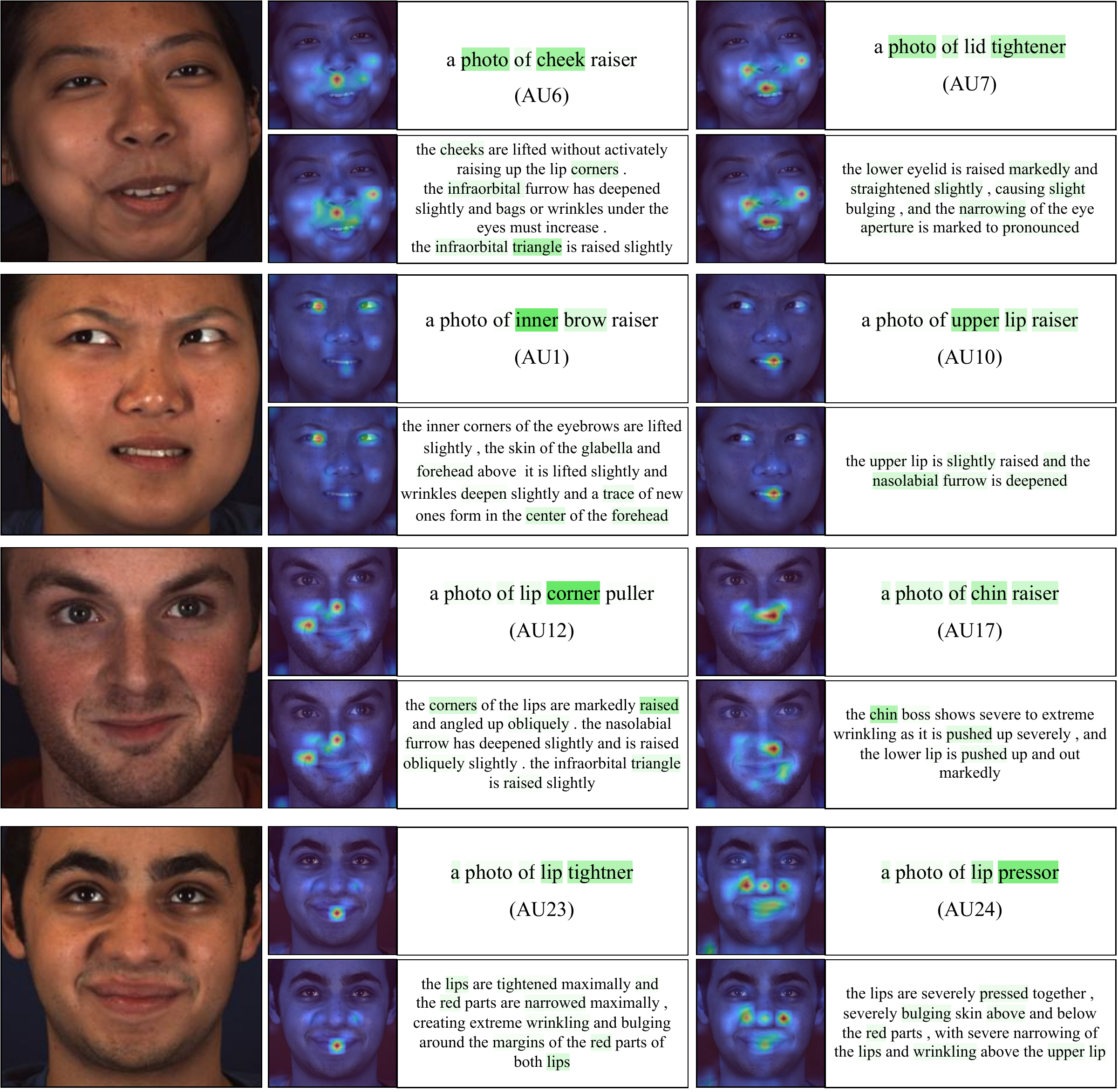}
    \caption{Heatmap samples of 8 AUs on BP4D. }
    \label{fig:heatmap_aur}
  \end{subfigure}
\vspace*{-.5em}
  \caption{Visualization of the relevancy heatmap between image-name and image-description pairs using GAE~\cite{Chefer_2021_ICCV}. 
  In (b), the (AU id) just indicates which AU it is, but not in the textual label name.}
  \label{fig:heatmap}
\end{figure*}

We also visualize the relevancy between the image and corresponding text queries by GAE~\cite{Chefer_2021_ICCV} in Figure~\ref{fig:heatmap}.
The image heatmap is arranged in increasing order of relevance from blue to red, while the text heatmap is arranged by increasing green intensity.
Examples of 8 expressions on AffectNet and 8 AUs on BP4D can be seen in Figure~\ref{fig:heatmap_fer} and Figure~\ref{fig:heatmap_aur}, respectively. 
% Gradients are calculated in the output of the first LayerNorm of the last Transformer block for both the image encoder and text encoder.
The text heatmap shows attention to relevant semantic words in the text, while the image heatmap localizes the corresponding regions of the face by querying the label name or label description.
We observe that the same face regions are highlighted when querying for label names and label descriptions, indicating that the text encoder has successfully learned to extract semantic knowledge even from the label names.

%------------------------------- Conclusion ------------------------------------------

\section{Discussion}

\B{Advanced Paring Method.} Unlike widely used object detection databases, which typically contain thousands of categories with distinct identities, facial behavior databases have limitations in terms of both the number of expression categories and identities, rendering traditional paring methods less efficient.
Pairing in CLEF is activity-based, where each activity is deliberately designed to elicit a specific expression, resulting in images with expression intensity, ranging from none to onset, peak, and offset.
Hence, the probability of grouping similar expressions is higher than self-supervised pairing (only the anchor itself is positive).

\B{Easy Extension.}
Using texts as label names facilitates easy extension with other information.
For example, intensity details can be integrated into label names by including phrases, ``with low intensity'', or ``with high intensity''.

\B{Limitation.} While our pre-trained CLEF can improve the performance of downstream tasks on various databases, it has certain limitations. 
1) Our pre-training approach relies on prior knowledge of coarse-grained textual descriptions, which may not be available in some databases.
We plan to address this issue in future updates by generating coarse-grained text descriptions.
2) We use a fixed prompt template for label names, and a random template for label descriptions, where the prompting is not fully explored. 
3) Variations in performance across AUs can be caused by semantic descriptive writing. 
Thus, further investigation into description writing is necessary.

\section{Conclusion}
This paper has proposed a weakly-supervised text-driven contrastive method that leverages the coarse-grained activity information to learn advanced facial representations.
The method minimizes intra-activity feature differences and maximizes inter-activity feature differences while disentangling the effects of subject identity features.
By incorporating textual label names and descriptions, the proposed network can directly be applied to FER and AUR tasks. 
CLEF achieves SOTA results on 3 widely used in-the-lab databases for AUR and 3 in-the-wild databases for FER. 
Ablation experiments show the effectiveness of weakly-supervised contrast learning in pre-training, as well as the validity of using textual information from activity, label name, and label description.
Compared to previous fine-grained pre-training methods, such as detecting landmarks, our coarse-grained approach requires less data processing while still achieving improvements.

%------------------------------- Acknowledgment ------------------------------------------

\section{Acknowledgment}
This work is supported by the NSF under grant CNS-1629898 and 
the Center of Imaging, Acoustics, and Perception Science (CIAPS) of the Research Foundation of Binghamton University.

{\small
\bibliographystyle{ieee_fullname}
\bibliography{egbib}
}

\end{document}

%%%%%%%%% TITLE
\title{Weakly-Supervised Text-driven Contrastive Learning for Facial Behavior Understanding \\
Supplementary Material}

\maketitle

%%%%%%%%% BODY TEXT

%------------------------------- Activity Descriptions ------------------------------------------
\section{Activity Descriptions}

Table~\ref{tab:activity_8} and ~\ref{tab:activity_10} show the activity descriptions of BP4D and BP4D+ respectively.

%%%%%%%%%%%%%%%%%%%%%%%   Activity Table BP4D ####################################

\begin{table}[th]
\caption{8 Activity descriptions the subjects participate in BP4D.}
\label{tab:activity_8}
\begin{center} %
\renewcommand{\arraystretch}{1.0}
\begin{tabular}{|l|m{7.0cm}|}
\hline
        & Activity Description   \\
\hline
A1       & Talk to the experimenter and listen to a joke (Interview). The target emotion is happiness or amusement \\ 
\hline
A2       & Watch and listen to a recorded documentary and discuss their reactions. The target emotion is sadness  \\
\hline
A3       & Experience sudden, unexpected burst of sound. The target emotion is surprise or startle   \\
\hline
A4       & Play a game in which they improvise a silly song. The target emotion is embarrassment \\
\hline
A5       & Anticipate and experience physical threat. The target emotion is fear or nervous \\
\hline
A6       & Submerge their hand in ice water for as long as possible. The target emotion is physical pain \\
\hline
A7       & Experience harsh insults from the experimenter. The target emotion is anger or upset \\
\hline
A8       & Experience an unpleasant smell. The target emotion is disgust \\
\hline
\end{tabular}
\end{center}
\end{table}

%%%%%%%%%%%%%%%%%%%%%%%   Activity Table BP4D+ ####################################

\begin{table}[th]
\caption{10 Activity descriptions the subjects participate in BP4D+.}
\label{tab:activity_10}
\begin{center} %
\renewcommand{\arraystretch}{1.0}
\begin{tabular}{|l|m{7.0cm}|}
\hline
        & Activity Description   \\
\hline
A1       & Interview: Listen to a funny joke. The target emotion is happiness or amusement \\ 
\hline
A2       & Graphic show: Watch 3D avatar of participant. The target emotion is surprise  \\
\hline
A3       & Video clip: 911 emergency phone call. The target emotion is sadness   \\
\hline
A4       & Experience a sudden burst of sound. The target emotion is startle or surprise \\
\hline
A5       & Interview: True or false question. The target emotion is skeptical \\
\hline
A6       & Improvise a silly song. The target emotion is embarrassment \\
\hline
A7       & Experience physical threat in dart game. The target emotion is fear or nervous \\
\hline
A8       & Cold pressor: Submerge hand into ice water. The target emotion is physical pain \\
\hline
A9       & Interview: Complained for a poor performance. The target emotion is anger or upset\\
\hline
A10      & Experience smelly odor. The target emotion is disgus \\
\hline
\end{tabular}
\end{center}
\end{table}

%------------------------------- Activity Descriptions ------------------------------------------

\section{Label Semantic Descriptions}

\subsection{Facial Expression}
Inspired by the work of SEV~\cite{sev}, we summarized 8 facial expression semantic descriptions based on the previous psychology study~\cite{ekman1971,matsumoto1992}.\\
Following descriptions are in \B{label name} : \B{label description} pattern.
\\\\
\noindent\B{Anger}: The eyebrows are lowered and pulled closer together, and the eyelids become squinted or raised. The lips would tighten or curl inwards, the corners of the mouth would point downwards, and the Jaw is tense and might jut forward slightly.\\
\noindent\B{Contempt}: The eyes would be unengaged, one side of the mouth is pulled up and back. One eyebrow may pull upwards and the head may tilt back slightly, making the gaze follow down the nose.\\
\noindent\B{Disgust}: The eyebrows are pulled down, and the nose is wrinkled. The upper lip is pulled up and the lips are loose. The eyes are narrow, the teeth may be exposed, and the cheeks may be raised.\\
\noindent\B{Fear}: The eyebrows are pulled up and together, and the upper eyelids are pulled up, and the lower eyelids are tense and drawn up. The mouth are stretched and drawn back, possibly exposing teeth. Vertical wrinkles may appear between the eyebrows.\\
\noindent\B{Happiness}: The eyes squint slightly, wrinkles appear at the corners of the eyes and the cheeks raise. The corners of the mouth move up at a diagonal, widening the mouth and the mouth may part, exposing teeth.\\
\noindent\B{Neutral}: The mouth is straight lined, the eyes are unfocused and the cheeks are slack. Not arch the eybrows, frown, smile or grimace. \\
\noindent\B{Sadness}: The eyebrows are lower and pulled closer together, and the inner corners of the eyebrows are angled up. The corners of the mouth are drawn downwards, and the lips may be either drawn in tightly or pouting outwards.\\
\noindent\B{Surprise}: The eyebrows are raised, and horizontal wrinkles would appear on the forehead. The jaw would go slack, the mouth would hang open loosely and the eyes would widen.\\

\subsection{Facial Action Unit}
The descriptions of AUs is written in SEV~\cite{sev}, which is based on the psychology study~\cite{ekman1997}.
We then has slightly modified these descriptions, which are shown in a pattern:\\
\B{AU id.} \B{label name} : \B{label description}.
\\

\noindent\B{AU1. inner brow raiser}: The inner corners of the eyebrows are lifted slightly, the skin of the glabella and forehead above it is lifted slightly and wrinkles deepen slightly and a trace of new ones form in the center of the forehead.

\noindent\B{AU2. outer brow raiser}: The outer part of the eyebrow raise is pronounced. The wrinkling above the right outer eyebrow has increased markedly, and the wrinkling on the left is pronounced. Increased exposure of the eye cover fold and skin is pronounced.

\noindent\B{AU4. brow lowerer}: The vertical wrinkles appear in the glabella and the eyebrows are pulled together. The inner parts of the eyebrows are pulled down a trace on the right and slightly on the left with traces of wrinkling at the corners.

\noindent\B{AU6. cheek raiser}: The cheeks are lifted without activately raising up the lip corners. The infraorbital furrow has deepened slightly and bags or wrinkles under the eyes must increase. The infraorbital triangle is raised slightly.

\noindent\B{AU7. lid tightener}: The lower eyelid is raised markedly and straightened slightly, causing slight bulging, and the narrowing of the eye aperture is marked to pronounced.

\noindent\B{AU9. nose wrinkler}: The nose is Wrinkled, the skin on bridge of the nose is drawn upwards, the nasal wings are lifted up, the infraorbital triangle is severely raised, and the upper part of the nasolabial fold is extremely deepened as the upper lip is drawn up slightly.

\noindent\B{AU10. upper lip raiser}: The upper lip is slightly raised and the nasolabial furrow is deepened.
    
\noindent\B{AU12. lip corner puller}: The corners of the lips are markedly raised and angled up obliquely. The nasolabial furrow has deepened slightly and is raised obliquely slightly. The infraorbital triangle is raised slightly.
	
\noindent\B{AU14. dimpler}: The lip corners are extremely tightened, and the wrinkling as skin is pulled inwards around the lip corners is severe. The skin on the chin and lower lip is stretched towards the lip corners, and the lips are stretched and flattened against the teeth.

\noindent\B{AU15. lip corner depressor}: The lip corners are pulled down slightly, with some lateral pulling and angling down of the corners, and slight bulges and wrinkles appear beyond the lip corners. 

\noindent\B{AU17. chin raiser}: The chin boss shows severe to extreme wrinkling as it is pushed up severely, and the lower lip is pushed up and out markedly.

\noindent\B{AU23. lip tightener}: The lips are tightened maximally and the red parts are narrowed maximally, creating extreme wrinkling and bulging around the margins of the red parts of both lips.

\noindent\B{AU24. lip pressor}: The lips are severely pressed together, severely bulging skin above and below the red parts, with severe narrowing of the lips and wrinkling above the upper lip.

\noindent\B{AU25. lips part}: The teeth is clearly shown, and the lips are separated slightly. Nothing suggests that the jaw has dropped even though the upper teeth are not clearly visible.

\noindent\B{AU26. jaw drop}: The jaw is lowered about as much as it can drop from relaxing of the muscles. The lips are parted to about the extent that the jaw lowering can produce.

%------------------------------- algorithm Descriptions ------------------------------------------

%------------------------------- algorithm pre-train ------------------------------------------

\begin{algorithm}[t]
\SetAlgoLined
\footnotesize
\begin{lstlisting}[language=python]
# encode_image: vision transformer 
# encode_text: text transformer
# img1,img2: image inputs of two augmentation
# activity: activity text
# t1, t2: two learned temperature parameters
# targets: activity labels

# extract feature representations for image
i_f1 = encode_image(img1)
i_f1 = i_f1/i_f1.norm(dim=1, keepdim=True)
i_f2 = encode_image(img2)
i_f2 = i_f2/i_f2.norm(dim=1, keepdim=True)
# extract feature representations for 
# activity description
a_f = encode_text(activity)
a_f = t_f/t_f.norm(dim=1, keepdim=True)
f_ii = torch.cat((i_f1, i_f2), 0)
f_ia = torch.cat((i_f1, a_f), 0)
# scaled cosine similarities
logit_ii = t1.exp()*i_f1 @ f_ii.t()
logit_it = t2.exp()*i_f1 @ f_ia.t()
# supervised contrastive loss function
loss_ii = sup_con_loss(logit_ii, targets) 
loss_ia = sup_con_loss(logit_it, targets) 
loss = (loss_ii + loss_ia)/2.0
\end{lstlisting}
\small
\caption{PyTorch-style pseudocode for CLEF in Pre-training}
\label{algo:pretrain}
\end{algorithm}
%------------------------------- algorithm fine-tune ------------------------------------------

\begin{algorithm}[th]
\footnotesize
\begin{lstlisting}[language=python]
# encode_image: Vision Transformer 
# encode_text: Text Transformer
# img: image input
# n_text: label name text
# d_text: label description text
# t1: learned temperature parameter
# t2: learned temperature parameter
# lambda: fixed hyperparameter 
# targets: facial expression or AU label

# extract feature representations for image
i_f = encode_image(img)
i_f = i_f/i_f.norm(dim=1, keepdim=True)
# extract feature representations for
# label name text
n_f = encode_text(n_text)
n_f = n_f/n_f.norm(dim=1, keepdim=True)
# extract feature representations for
# description text
d_f = encode_text(d_text)
d_f = d_f/d_f.norm(dim=1, keepdim=True)
# scaled cosine similarities
logit_in = t1.exp()*i_f @ n_f.t()
logit_dn = t2.exp()*d_f @ n_f.t()
# loss function
# if task is FER, task_loss: cross_entropy_loss
# if task is AUR, task_loss: bce_loss
loss_in = task_loss(logit_in, targets)
labels = torch.arange(n_text.shape[0])
loss_dn = cross_entropy_loss(logit_dn, labels) 
loss = (lambda * loss_in + loss_dn)/2.0
\end{lstlisting}
\small
\caption{PyTorch-style pseudocode for CLEF in Fine-tuning}
\label{algo:finetune}
\end{algorithm}

\begin{table*}[th]
\caption{Fine-tuning Settings}
\label{tab:setting_fine}
\centering %
\renewcommand{\arraystretch}{1.0}
\begin{tabular}{l|*{5}c}
\toprule
Database    & epochs    & lr    & \S{Warm-up epochs}& \S{lr schedule}     & \S{weight decay}  \\
\toprule
BP4D        & 3         & 0.0002& 1                 & cosine decay: [1, 3]  & 0.01                \\  
BP4D+       & 3         & 0.0002& 1                 & cosine decay: [1, 3]  & 0.01                  \\
DISFA       & 5         & 0.0001& 0                 & steps: [2:0.1, 5:0.5] & 0.01              \\ 
Affect-Net  & 3         & 0.0002& 1                 & cosine decay: [1, 3]  & 0.01             \\
RAF-DB      & 5         & 0.0002& 1                 & cosine decay: [1, 5]  & 0.01             \\
FER+        & 7         & 0.0001& 1                 & cosine decay: [3, 7]  & 0.05           \\
\bottomrule

\end{tabular}
\end{table*}

\section{Pseudo-codes}

We provide the pytorch-style pseudo-codes for both pretraining and finetuning in Algorithm~\ref{algo:pretrain} and ~\ref{algo:finetune}.

\section{Text Prompt templates}
Let N denotes label name, D indicates label descriptions, and A represents activity descriptions.
For label name prompting, only one template is used, i.e., ``a photo of a person with \{N\}.''.
Label description prompting is randomly chose from one of the AU or expression templates.

\noindent \B{AU Label Description Templates}:
\begin{itemize}
  \item ``a photo of a person with \{D\}.''
  \item ``a photo shows a person that \{D\}.''
  \item ``a photo of one has \{D\}.''
  \item ``a photo of a person that \{D\}.''
  \item ``a photo of a face with \{D\}.''
  \item ``a photo of a person has \{D\}.''
  \item ``a good photo of a person that \{D\}.''
  \item ``the photo of a face that \{D\}.''
  \item ``the photo of a person that \{D\}.''
  \item ``a photo of a face where \{D\}.''
  \item ``a photo shows facial action unit that \{D\}.''
  \item ``a cropped photo of face that \{D\}.''
  \item ``a clean photo of a person that \{D\}.''
  \item ``a facial action unit where \{D\}.''
\end{itemize}

\noindent \B{Expression Label Description Templates}:
\begin{itemize}
  \item ``a photo of a person with \{D\}.''
  \item ``a photo shows a person with \{D\}.''
  \item ``a photo of one has \{D\}.''
  \item ``a photo of a face that \{D\}.''
  \item ``a photo of a person has \{D\}.''
  \item ``a good photo of a person in \{D\}.''
  \item ``the photo of a face in \{D\}.''
  \item ``a cropped photo of face that \{D\}.''
  \item ``a clean photo of a person with \{D\}.''
  \item ``a facial expression where \{D\}.''
  \item ``a photo of facial expression that \{D\}.''
\end{itemize}

\noindent Activity description prompting is randomly chose from one of the following templates.

\noindent \B{Activity Description Templates}:
\begin{itemize}
  \item ``a photo of a person from an activity that \{A\}.''
  \item ``a photo shows a person in the activity that \{A\}.''
  \item ``a photo of an activity that \{A\}.''
  \item ``a photo of a person participated in an activity that \{A\}.''
  \item ``a photo of a face from the activity that \{A\}.''
  \item ``a photo of a person was in an activity that \{A\}.''
  \item ``a good photo of the activity where \{A\}.''
  \item ``a photo of a person joined in an activity that \{A\}.''
  \item ``a good photo of a person in an activity that \{A\}.''
  \item ``a cropped photo of face from an activity where \{A\}.''
  \item ``a clean photo of a person in the activity that \{A\}.''
  \item ``an activity where \{A\}.''

\end{itemize}

\section{More Implementation Details}

Table~\ref{tab:setting_fine} and ~\ref{tab:setting_pre} show the detail implementation settings for fine-tuning and pre-training respectively. 
The settings not shown in Table~\ref{tab:setting_fine} are the same as the pre-training settings.
Note that only augmentation 1 is applied in the fine-tuning image augmentation.

\begin{table}[ht]
\caption{Pre-training Settings}
\label{tab:setting_pre}
\begin{center} %
\renewcommand{\arraystretch}{1.0}
\begin{tabular}{p{4.0cm}|c}
\toprule
config                  & value   \\
\toprule
Batch size              & 64\\
Vocabulary size         & 49408\\
Training epochs         & 5\\
Warm-up epochs          & 1\\
learning rate schedule  & cosine decay\\
learning rate           & $10^{-5}$\\
min learning rate       & $10^{-6}$\\
weight decay            & 0.01\\
AdamW betas             & (0.9, 0.999)\\
augmentation 1          & HorizontalFlip\\
augmentation 2          & ResizedCrop\\
                        & HorizontalFlip\\
                        & RandomRotation\\
\bottomrule
\end{tabular}
\end{center}
\vspace{-10pt}
\end{table}

\section{More Ablation study}

\begin{figure}[ht]
  \centering
  \includegraphics[width=0.47\textwidth]{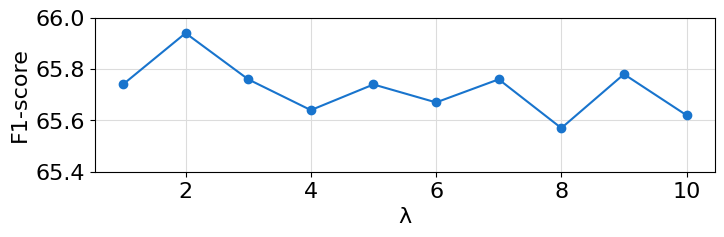}
  \caption{F1-score with different $\lambda$ on BP4D}
  \label{fig:lambada}
\end{figure}

\B{Evaluation of different $\lambda$.}
In this section, we evaluate the performance on BP4D by setting different hyperparameters $\lambda$, which can be seen in Firuge~\ref{fig:lambada}
The performance reaches its peak when $\lambda$ is set to 2, which is attributed to the fact that loss from Image-Name pairs plays a major role in back propagation as Image-Name pairs are more diverse than Name-Description pairs.

\section{More Visualization}

\begin{figure}[th]
  \centering
  \begin{subfigure}{0.98\linewidth}
    \includegraphics[width=\textwidth]{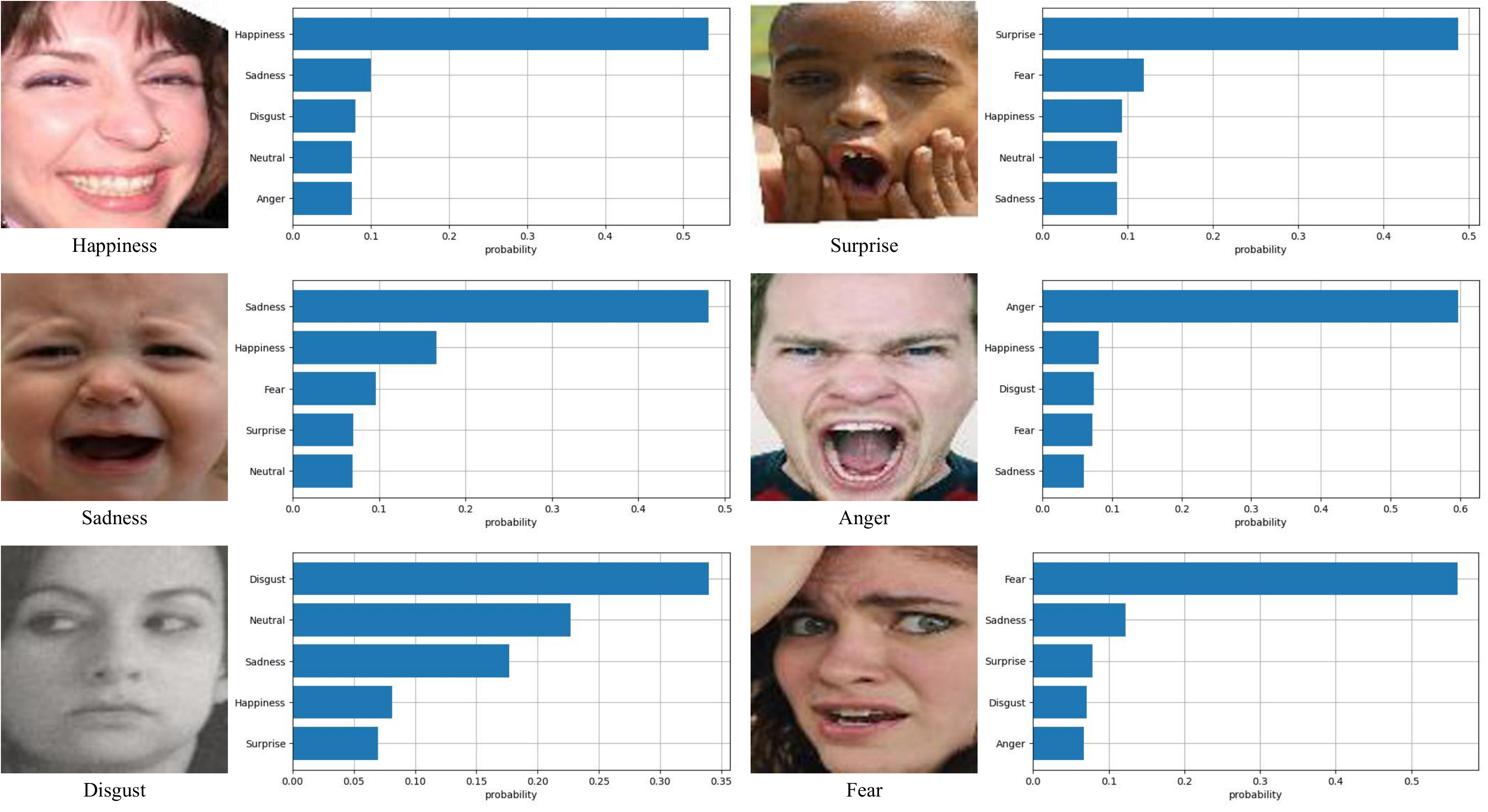}
    \caption{successful prediction examples}
  \end{subfigure}
  \begin{subfigure}{0.98\linewidth}
    \includegraphics[width=\textwidth]{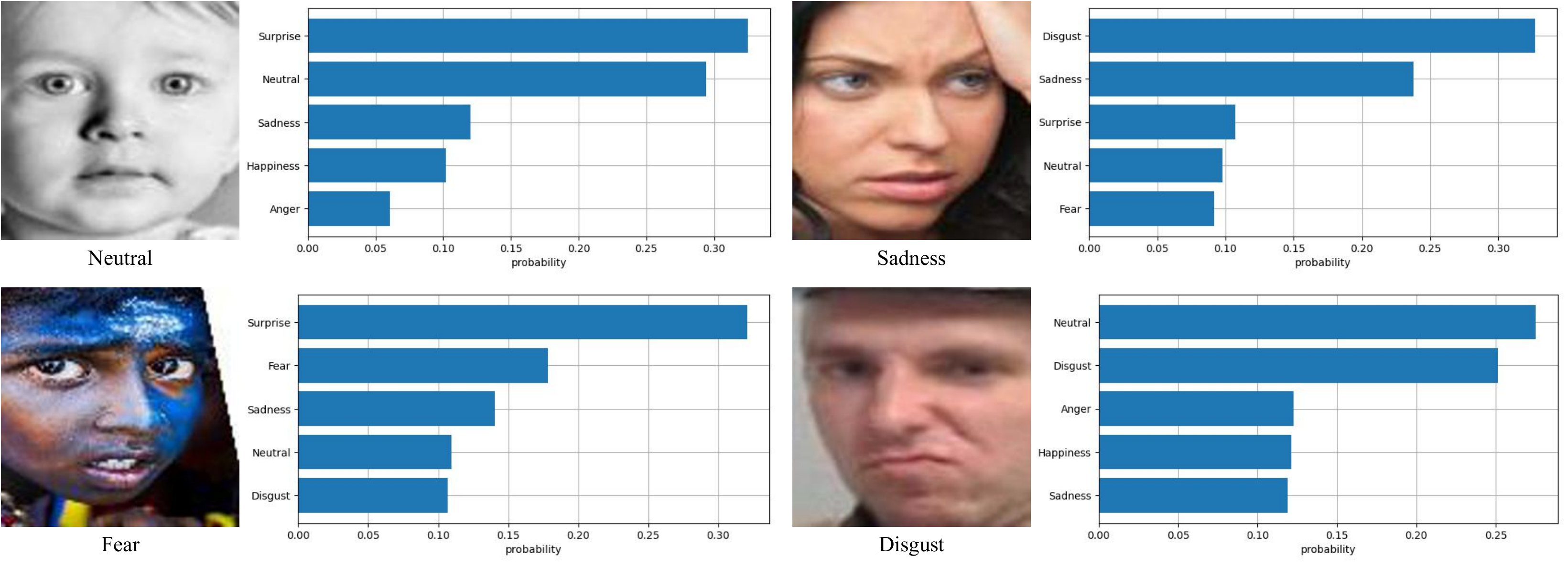}
    \caption{fail prediction examples}
  \end{subfigure}
    \caption{Visualization of image samples and the probabilities of their top 5 predictions on RAF-DB. 
    The query texts are in the template of ``a photo of a person with \{N\}''}
    \label{fig:figure_prob}
\end{figure}

Figure~\ref{fig:figure_prob} shows more visualizations of prediction probability on RAF-DB.
The query text is in ``a photo of a person with \{N\}'' format.
Both success and failure examples are shown in it.

%%%%%%%%% REFERENCES
{\small
\bibliographystyle{ieee_fullname}
\bibliography{egbib}
}